\documentclass[journal]{IEEEtran}
\usepackage{booktabs}
\usepackage{multirow}

\usepackage{amsmath}
\usepackage{amssymb}
\usepackage{algorithm}
\usepackage[noend]{algpseudocode}
\usepackage{graphicx}
\usepackage{subcaption}
\usepackage[most]{tcolorbox} 
\usepackage{multirow}
\begin{document}

\title{Extracting Actionable Insights from Building Energy Data using Vision LLMs on Wavelet and 3D Recurrence Representations}

\author{\IEEEauthorblockN{
Amine Bechar\IEEEauthorrefmark{3}, 
Adel Oulefki\IEEEauthorrefmark{4}
Abbes Amira\IEEEauthorrefmark{3},
Fatih Kurogollu\IEEEauthorrefmark{3} and
Yassine Himeur\IEEEauthorrefmark{2}
}\\
\IEEEauthorblockA{\IEEEauthorrefmark{3}Department of Computer Science, University of Sharjah, Sharjah, UAE
(abechar@sharjah.ac.ae, aaamira@sharjah.ac.ae, fkurogollu@sharjah.ac.ae)}\\
\IEEEauthorblockA{\IEEEauthorrefmark{4}Research Institut of Science and Technology (RISE), University of Sharjah, UAE
(aoulefki@sharjah.ac.ae)}\\
\IEEEauthorblockA{\IEEEauthorrefmark{2}College of Engineering and Information Technology, University of Dubai Dubai, UAE (yhimeur@ud.ac.ae)}\\}

\markboth{IEEE International Conference on Data Mining, June~2025}%
{}

\maketitle

\begin{abstract}

The analysis of complex building time-series for actionable insights and recommendations remains challenging due to the nonlinear and multi-scale characteristics of energy data. To address this, we propose a framework that fine-tunes visual language large models (VLLMs) on 3D graphical representations of the data. The approach converts 1D time-series into 3D representations using continuous wavelet transforms (CWTs) and recurrence plots (RPs), which capture temporal dynamics and localize frequency anomalies. These 3D encodings enable VLLMs to visually interpret energy-consumption patterns, detect anomalies, and provide recommendations for energy efficiency. We demonstrate the framework on real-world building-energy datasets, where fine-tuned VLLMs successfully monitor building states, identify recurring anomalies, and generate optimization recommendations. Quantitatively, the Idefics-7B VLLM achieves validation losses of 0.0952 with CWTs and 0.1064 with RPs on the University of Sharjah energy dataset, outperforming direct fine-tuning on raw time-series data (0.1176) for anomaly detection. This work bridges time-series analysis and visualization, providing a scalable and interpretable framework for energy analytics.

\end{abstract}


\IEEEpeerreviewmaketitle

\section{Introduction}

\IEEEPARstart{T}{he} drive toward energy efficiency in buildings is a cornerstone of sustainable development, with significant economic, environmental, and social implications \cite{papadakis2023review}. Buildings are responsible for a substantial share of global energy consumption and greenhouse gas emissions, and 36\% of total emissions in Europe alone, making them a critical focus for climate action and resource conservation \cite{fathi2024sustainability}. Improving energy performance in both new and existing structures not only reduces operational costs but also supports larger sustainability goals by improving the well-being of the occupants \cite{oulefki2025innovative, hafez2022energy}. 

Time-series data on building energy consumption play a pivotal role in understanding operational dynamics and uncovering inefficiencies \cite{oulefki2024dataset, liu2022data, pickering2018building}. By analyzing patterns in electricity, heating, and cooling usage, stakeholders can identify opportunities for targeted interventions, optimize building systems, and ultimately drive more efficient energy use \cite{hauashdh2024integrated}. Advanced data-driven approaches, including machine learning and smart computing, are increasingly being used to classify, predict, and interpret trends in energy consumption, enabling better decision-making and continuous improvement in building performance \cite{ahmad2022data}. The analysis of building time series is difficult because of its complex nonlinear patterns and scale-related properties, such as multiple frequencies \cite{althelaya2021combining}. 

Traditional time series analysis methods face significant constraints when used for data on building energy. Statistical models and classical machine learning techniques are struggling to capture the complex nonlinearities and multiscale dynamics that characterize energy consumption patterns \cite{huynh2023knowledge}. These approaches often lack clarity, which prevents them from providing actionable recommendations for optimization. Additionally, scalability becomes problematic with high-dimensional or long-sequence data, creating computational challenges \cite{jiang2025large}. The continuing gap between raw data analysis and the generation of human-readable insights leaves energy managers without intuitive and practical guidance on how to improve energy efficiency \cite{li2025review}. Energy efficiency for time-series analysis commonly employs techniques such as LSTMs, Prophet, and gradient boosting algorithms (XGBoost, CatBoost, LightGBM) for consumption forecasting, alongside support vector regression (SVR), which excels in non-linear data \cite{gopal2025advanced, thakur2023predictive, liu2019forecasting}. 

Large language models (LLMs) are revolutionizing the field of energy efficiency in data-driven environments, bringing transformative potential to the mining, analysis, and optimization of energy usage patterns \cite{zhang2024advancing}. These advancements are made possible by integrating LLMs into the control and management layers of energy systems through artificial intelligence (AI) agents \cite{LIU2024151625}. One of the key strengths of LLMs lies in their ability to process and analyze diverse data types using advanced data mining techniques \cite{wan2024tnt}. This improves the accuracy and efficiency of identifying energy savings opportunities, particularly in scenarios where traditional methods struggle due to high complexity and large volumes of data. LLMs can automate a wide range of tasks, from monitoring energy consumption patterns \cite{xiao2024exploring} to recommending optimization strategies \cite{oprea2024recommendation}, thus improving overall system efficiency and resilience.

To address feature extraction challenges, transformations such as continuous wavelet transforms (CWT) and recurrence plots (RP) convert 1D data into 3D representations, capturing time-frequency dynamics and temporal patterns \cite{zhanlong2021research, jiang2022time}. Furthermore, vision-language models (VLMs) like CLIP and DALL-E have revolutionized multimodal understanding. For example, demonstrating strong performance in biomedical imaging and autonomous driving \cite{xing2024survey, zhang2025vl}. However, a critical gap remains: no framework uses the visual interpretation capabilities of VLLMs to analyze transformed 3D data (CWT/RPs) to produce actionable, domain-specific insights. Current VLLM applications offer a broad understanding, but lack tailored mechanisms to convert building energy visualizations into interpretable anomaly detection and optimization recommendations \cite{chen2023collaborative}.

A novel framework has been introduced that fine-tunes VLLMs to transform building energy analytics. The core innovation lies in converting 1D time-series data into interpretable 3D visual representations, specifically CWTs and RPs. CWTs encode time-frequency dynamics to localize transient anomalies \cite{zhang2019wavelet}, while RPs capture temporal patterns \cite{marwan2007recurrence}. These transformations enable VLLMs to "see" energy data as visual scenes, harnessing their multimodal reasoning capabilities to interpret complex consumption patterns across multiple scales, detect anomalies through visual feature recognition, and generate natural language optimization recommendations. By bridging raw sensor data, this framework directly converts spectral and temporal data into actionable energy intelligence and provides recommendations. This paper proposes several contributions.
\begin{itemize}
    \item Proposing a pipeline that converts 1D building energy time series into 3D visual representations using CWTs and RPs. This transformation encodes frequency-localized anomalies via CWTs and multiscale temporal dynamics via RPs, enabling VLLMs to process energy data as visual scenes. By using pre-trained multimodal VLLMs' capabilities, the framework overcomes limitations of traditional time-series analysis methods that struggle with nonlinear patterns and scalability.

    \item Demonstrating practical efficacy by deploying the framework on energy datasets from the University of Sharjah (UoS) buildings. This validates the approach in noisy and diverse environments typical of operational environments and demonstrates its robustness in the field beyond theoretical simulations.

    \item  Integrating three critical energy management functions: (i) Monitoring by interpreting multiscale consumption patterns; (ii) Anomaly Detection through visual feature recognition; and (iii) Natural Language Recommendations generating actionable optimization advice.
    
    \item Providing quantitative evidence (validation loss) that finetuning VLLMs on 3D representations outperforms direct time-series finetuning for anomaly detection to confirm the advantage of visual encodings in capturing complex patterns.
\end{itemize}
The remainder of this paper is structured as follows. Section II provides the background and related works, reviewing existing approaches to energy analytics. Section III presents the proposed framework, detailing the novel architecture that combines time series analysis with visual reasoning capabilities. Section IV covers experiments and results, evaluating the performance of the framework in multiple energy datasets and comparing metrics against baseline methods and approaches. Finally, Section V discusses the implications of the results, analyzes limitations and future research directions.

\section{Methodology}
\begin{figure*}[t]
        \begin{center}
        \includegraphics[width=420pt]{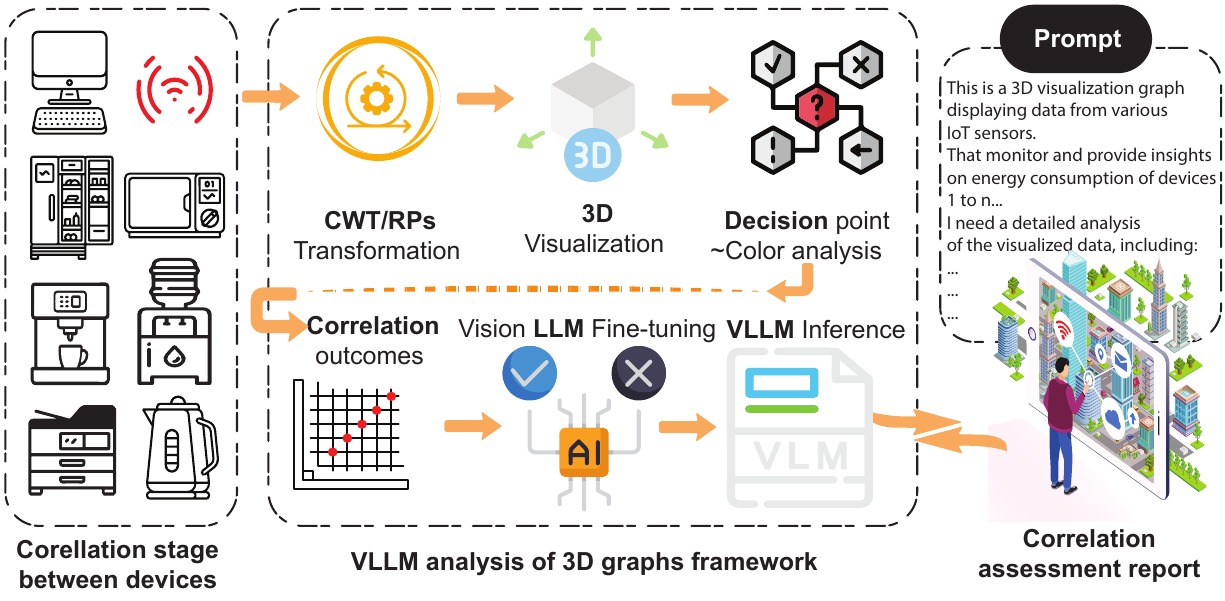}\\
        \end{center}
        \caption{Multi device energy efficiency correlation analysis using 3D CWT/RFs visualization and VLLM}
        \label{fig:cwt_process}
\end{figure*}

The proposed framework transforms 1D building energy time series into 3D visual representations for interpretation by VLLMs. Fig. \ref{fig:cwt_process} illustrates our transformation pipeline from raw sensor data to CWT representation, and then passes the graphs to the VLLM for fine-tuning. This section details the core components.

\subsection{Continuous Wavelet Transform (CWT)}
\label{subsec:cwt}

CWT \cite{aguiar2014continuous} serves as a fundamental transformation technique within our framework, converting 1D building energy time series data into interpretable 3D time-frequency representations.  This transformation is mathematically defined as:

\begin{equation}
C(a, b) = \frac{1}{\sqrt{|a|}} \int_{-\infty}^{\infty} x(t) \psi^*\left(\frac{t - b}{a}\right)  dt
\label{eq:cwt}
\end{equation}

where:
\begin{itemize}
    \item $x(t)$ is the original energy consumption time-series signal
    \item $\psi(t)$ denotes the mother wavelet function
    \item $a$ represents the scale parameter (inversely proportional to frequency)
    \item $b$ is the translation parameter (time shift)
    \item $\psi^*$ indicates the complex conjugate of the wavelet
\end{itemize}

CWT provides critical advantages for energy efficiency in buildings through several key capacities: It allows for multi-resolution analysis by simultaneously capturing both high-frequency transient events and low-frequency trends using scalable kernels, and also provides for anomaly localization by identifying frequency-localised anomalies by detecting abrupt coefficient changes at different scales. In our implementation, the complex Morlet wavelet is the main function, because of its optimal balance between time and frequency orientation, and it offers visual interpretation by producing 3D scallops (time frequency maps) as visual input to VLLMs.

\begin{equation}
\psi(t) = \pi^{-1/4} e^{i\omega_0 t} e^{-t^2/2}
\label{eq:morlet}
\end{equation}

where $\omega_0$ denotes the central frequency of the wavelet. The resulting scalogram $S(t, f)$ is computed as $S(t, f) = |C(a,b)|^2$ with $f = f_s/a$ ($f_s$: sampling frequency). This representation preserves both temporal dynamics and spectral characteristics, enabling the VLLM vision encoder to detect invisible patterns in the original 1D signal.

\begin{figure*}[t]
\centering
\setlength{\tabcolsep}{0pt} 
\begin{tabular}{c c c c}
\begin{minipage}{0.24\textwidth}
  \centering
  \includegraphics[width=0.95\linewidth]{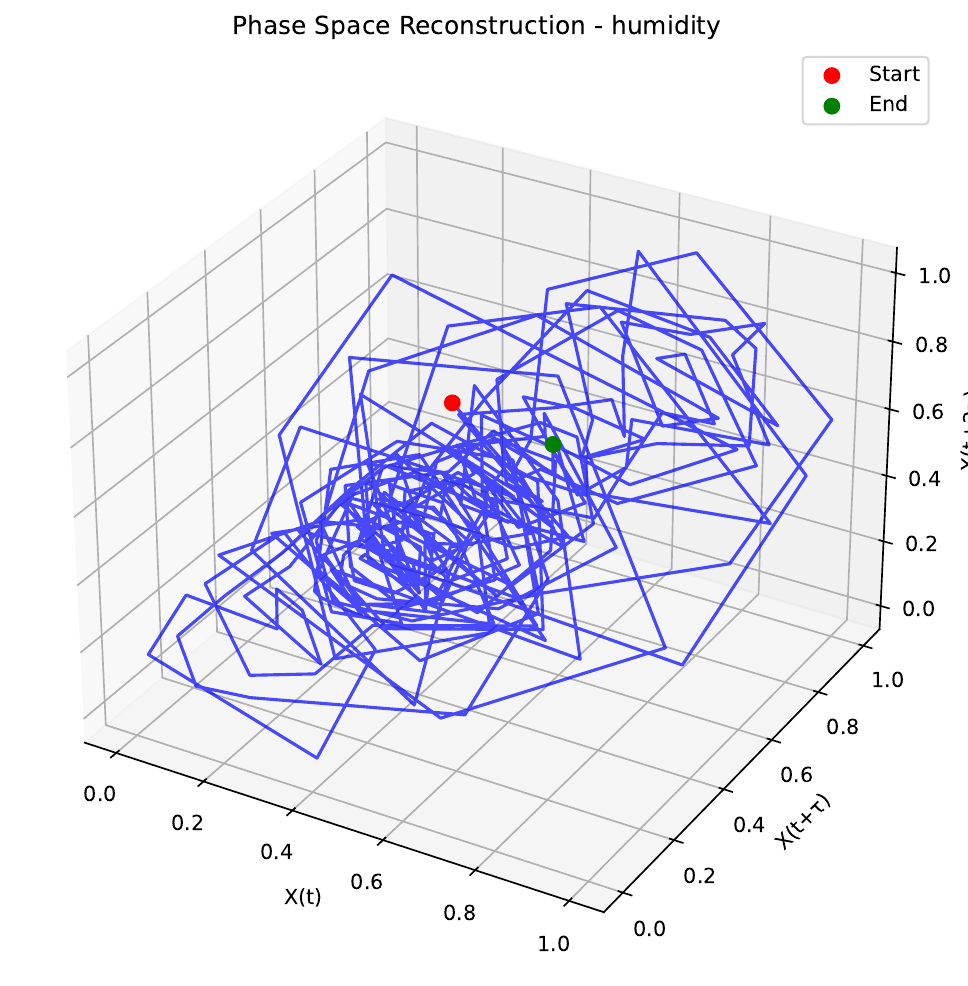}
  \subcaption{Humidity recurrence \\ pattern}
  \label{fig:sub1}
\end{minipage} &
\begin{minipage}{0.24\textwidth}
  \centering
  \includegraphics[width=0.95\linewidth]{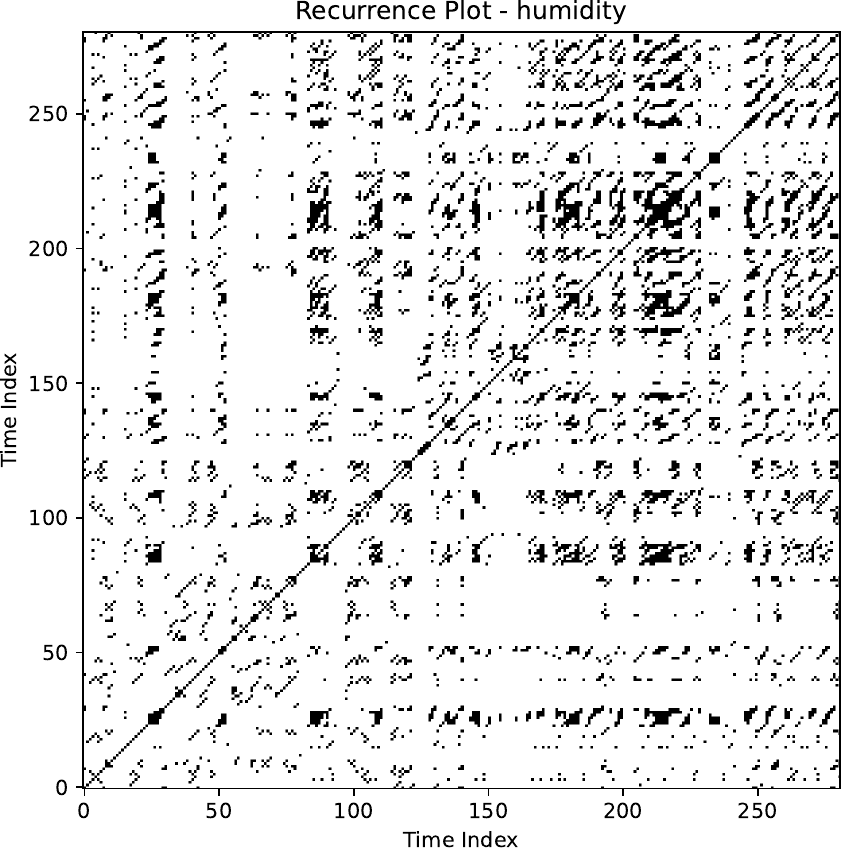}
  \subcaption{ Extended humidity \\ recurrence}
  \label{fig:sub2}
\end{minipage} &
\begin{minipage}{0.24\textwidth}
  \centering
  \includegraphics[width=0.95\linewidth]{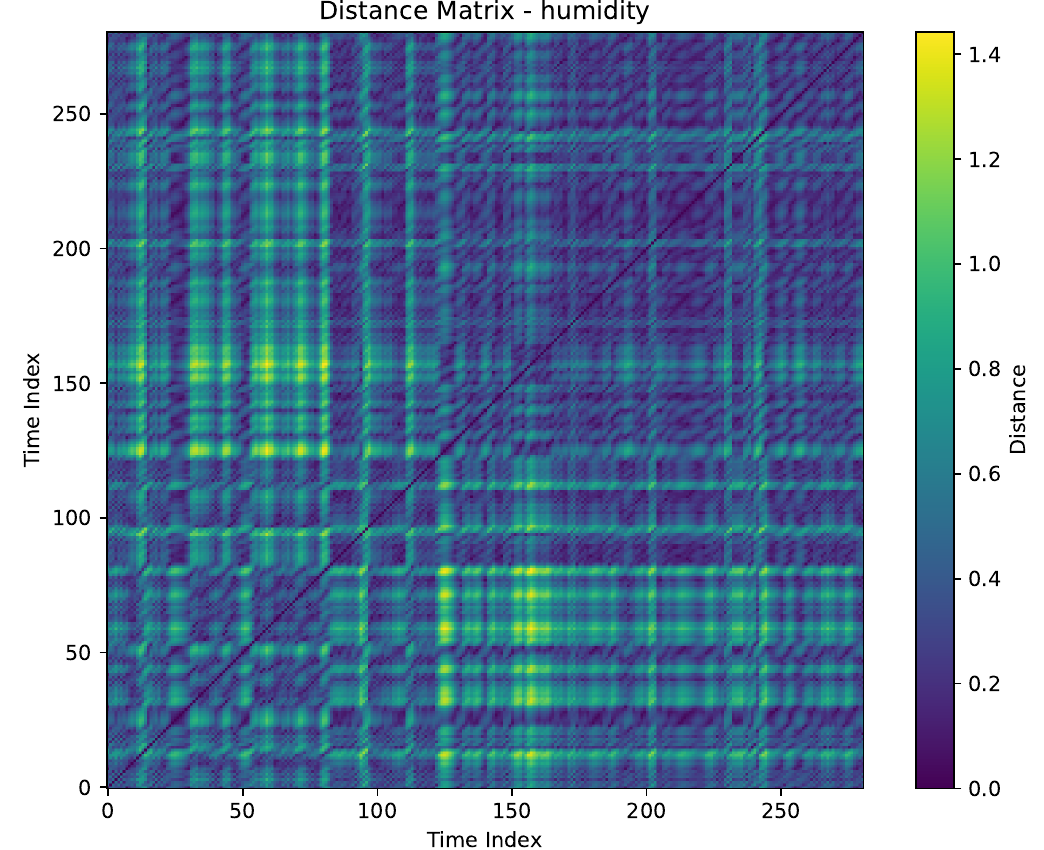}
  \subcaption{ Consumption scalogram}
  \label{fig:sub3}
\end{minipage} &
\begin{minipage}{0.24\textwidth}
  \centering
  \includegraphics[width=0.95\linewidth]{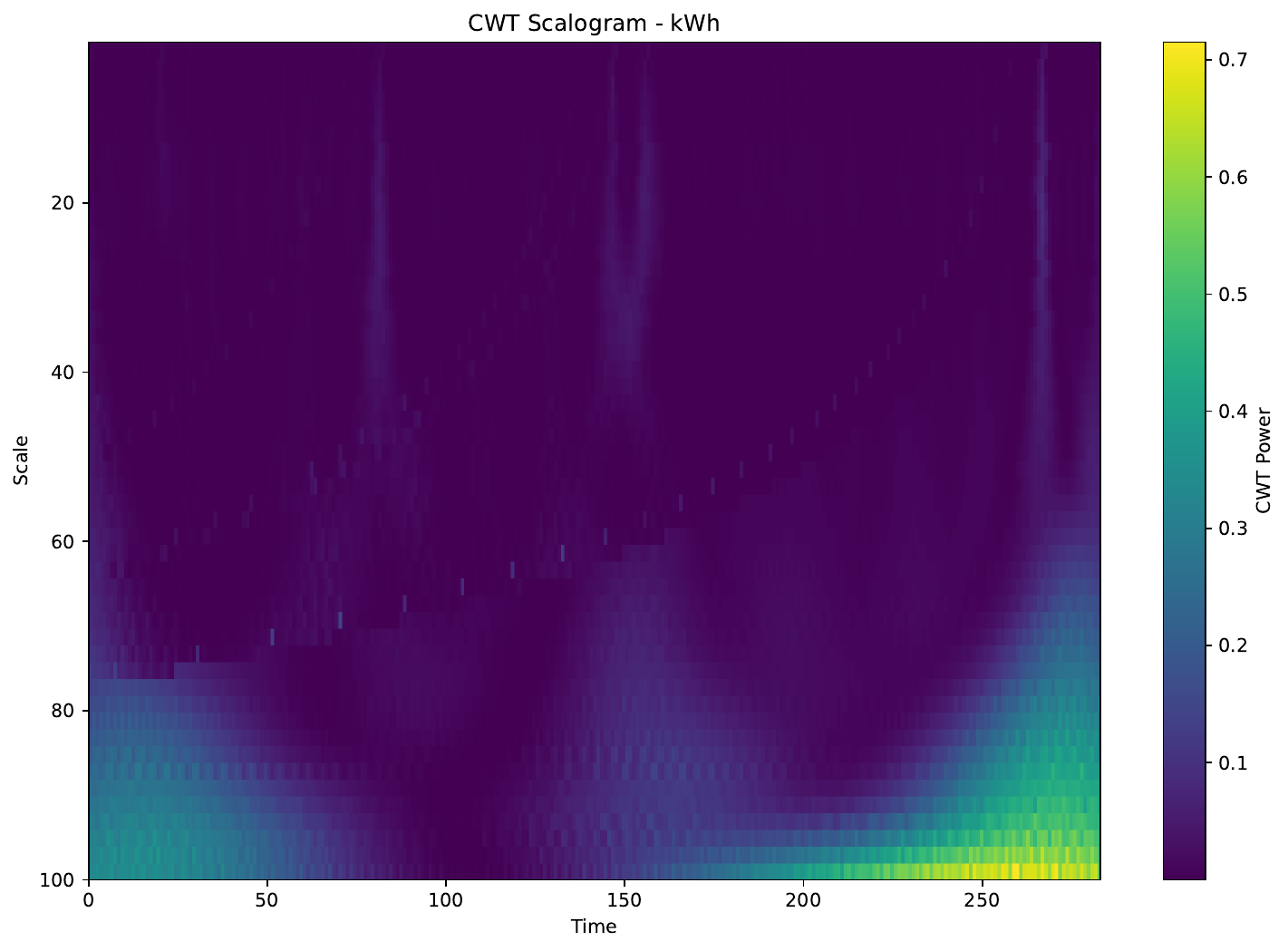}
  \subcaption{Humidity \& energy \\ analysis}
  \label{fig:sub4}
\end{minipage}
\end{tabular}
\caption{ A computational analysis results displaying recurrence plots, scalogram transformations, and comparative energy-humidity pattern recognition across multiple analytical dimensions}
\label{fig:test}
\end{figure*}

Each scalogram encodes approximately 24 hours of building energy consumption, with intensity values normalized to [0, 1] using min-max scaling. The resulting $H \times W$ pixel images form the primary visual input for anomaly detection tasks, where transient disturbances manifest themselves as localized intensity variations in specific frequency bands. Fig. \ref{fig:test} represents steps in creating a 3D analysis of humidity and energy patterns based on CWT and time series data.

\subsection{Model Architecture}

\subsubsection{Vision-Language Large Models (VLLMs) as Core Architecture}
The proposed framework takes advantage of VLLMs as its foundational architecture to interpret 3D visual representations of building energy time series data. This approach capitalizes on VLLMs' emerging multi-modal reasoning capabilities, which enable synergistic processing of heterogeneous data modalities. Transformation of temporal energy sequences into visual representations exploits the spatial pattern recognition strengths of vision transformers while bypassing limitations of sequential models in capturing long-range dependencies \cite{vaswani2017attention}. This architectural selection is justified by the demonstrated proficiency of VLLM in cross-modal alignment tasks, where visual characteristics must be contextualized within domain semantic frameworks \cite{alayrac2022flamingo}.

\subsubsection{VLLM Architectural Foundation}
VLLMs provides the baseline unified encoder-decoder architecture, formalized as:

\begin{equation}
\text{Output} = \mathcal{F}_{\theta}(\text{Image}, \mathcal{P}_{\tau}(\text{Task\_Prompt}, \text{Question}))
\end{equation}

where $\mathcal{F}{\theta}$ denotes the parameterized model and $\mathcal{P}{\tau}$ represents the prompt embedding function. Key components include:

\begin{enumerate}
\item \textbf{Vision Encoder}: A Vision Transformer (ViT) \cite{dosovitskiy2020image} with $L$ layers, patch size $P \times P$, and hidden dimension $D_v$. For input image $\mathbf{I} \in \mathbb{R}^{H \times W \times C}$, it produces patch embeddings:
\begin{equation}
\mathbf{Z}0 = [\mathbf{z}{\text{class}}; \mathbf{E}\mathbf{p}_1; \mathbf{E}\mathbf{p}2; \cdots; \mathbf{E}\mathbf{p}N] + \mathbf{E}{\text{pos}}
\end{equation}
where $N = HW/P^2$, $\mathbf{E} \in \mathbb{R}^{P^2C \times D_v}$ is the patch projection, and $\mathbf{E}{\text{pos}}$ the positional encoding. The encoder outputs latent features $\mathbf{V} \in \mathbb{R}^{N \times D_v}$.

\item \textbf{Language Decoder}: A $M$-layer transformer decoder that implements causal attention. Given tokenized text $\mathbf{T} \in \mathbb{R}^{S}$, it computes:
\begin{equation}
\mathbf{H}^l = \text{DecoderBlock}(\mathbf{H}^{l-1}, \mathbf{V}, \mathbf{M}{\text{mask}})
\end{equation}
where $\mathbf{M}{\text{mask}}$ prevents information leakage.

\item \textbf{Multimodal Fusion}: Cross-attention mechanisms bridge modalities:
\begin{equation}
\text{CrossAttn}(\mathbf{Q}, \mathbf{K}, \mathbf{V}) = \text{softmax}\left(\frac{\mathbf{Q}\mathbf{K}^T}{\sqrt{d_k}}\right)\mathbf{V}
\end{equation}
with $\mathbf{Q} = \mathbf{H}^{l-1}\mathbf{W}_q$ and $\mathbf{K},\mathbf{V}$ derived from visual features $\mathbf{V}$. This enables gradient-based feature alignment.
\end{enumerate}

\subsubsection{Domain-Specific Adaptation Strategy}
Task-specific conditioning is implemented through instruction templating:
\begin{equation}
\text{Input} = \underbrace{\texttt{<ANALYSIS\_TYPE>}}_{\text{Task Prompt}} \oplus \underbrace{\texttt{"Query: "} \mathcal{Q}}_{\text{Question}} \oplus \underbrace{\mathcal{I}}_{\text{Image}}
\end{equation}

where $\oplus$ denotes concatenation and $\mathcal{Q}$ is a query in natural language (e.g., ``Identify anomalies between 2023-07-01 and 2023-07-05''). The token $\texttt{<ANALYSIS\_TYPE>}$ is selected from $\{Monitoring, AnomalyDetection, Recommendation\}$ to configure the behavior of the decoder.

\subsubsection{Supervised Fine-Tuning Protocol}
The adaptation employs a two-stage optimization procedure:

\begin{enumerate}
\item \textbf{Selective Parameter Optimization}: The vision encoder parameters $\Theta_v$ remain frozen to preserve the pre-referenced visual representations. Only language decoder parameters $\Theta_d$ and cross-attention weights $\Theta_c$ are updated, preventing catastrophic forgetting while enabling domain-specific linguistic adaptation.

\item \textbf{Optimization Configuration}: 
The optimization setup employed several key components to ensure effective training. Gradient accumulation utilized 8-step accumulation ($\Gamma=8$) to simulate batch size $B_{\text{eff}} = \Gamma \times B$ while maintaining memory constraints. The learning rate schedule implemented a linear warmup over 50 steps to peak $\eta_{\text{max}} = 1 \times 10^{-4}$, followed by cosine decay according to:
\begin{align}
\eta_t &= \eta_{\text{min}} + \frac{1}{2}(\eta_{\text{max}} - \eta_{\text{min}}) \notag \\
&\quad \times \left(1 + \cos\left(\frac{t - t_{\text{warm}}}{t_{\text{max}} - t_{\text{warm}}} \pi\right)\right)
\end{align}
The loss function employed autoregressive cross-entropy minimization defined as:
\begin{align}
\mathcal{L} = -\frac{1}{NT} \sum_{i=1}^{N} \sum_{j=1}^{T} \sum_{k=1}^{|\mathcal{V}|} \, \mathbf{y}_{i,j,k} 
& \notag \\
\cdot \log P_\theta(&\hat{\mathbf{y}}_{i,j,k} \,|\, \mathbf{x}_i, \mathbf{y}_{i,<j})
\end{align}
where $|\mathcal{V}|$ is vocabulary size, $\mathbf{x}_i = (\mathcal{I}_i, \mathcal{P}_i, \mathcal{Q}_i)$, and $T$ is sequence length.
\end{enumerate}

\subsubsection{Model Configuration}
\begin{figure}[t]
    \begin{center}
    \includegraphics[width=0.42\textwidth]{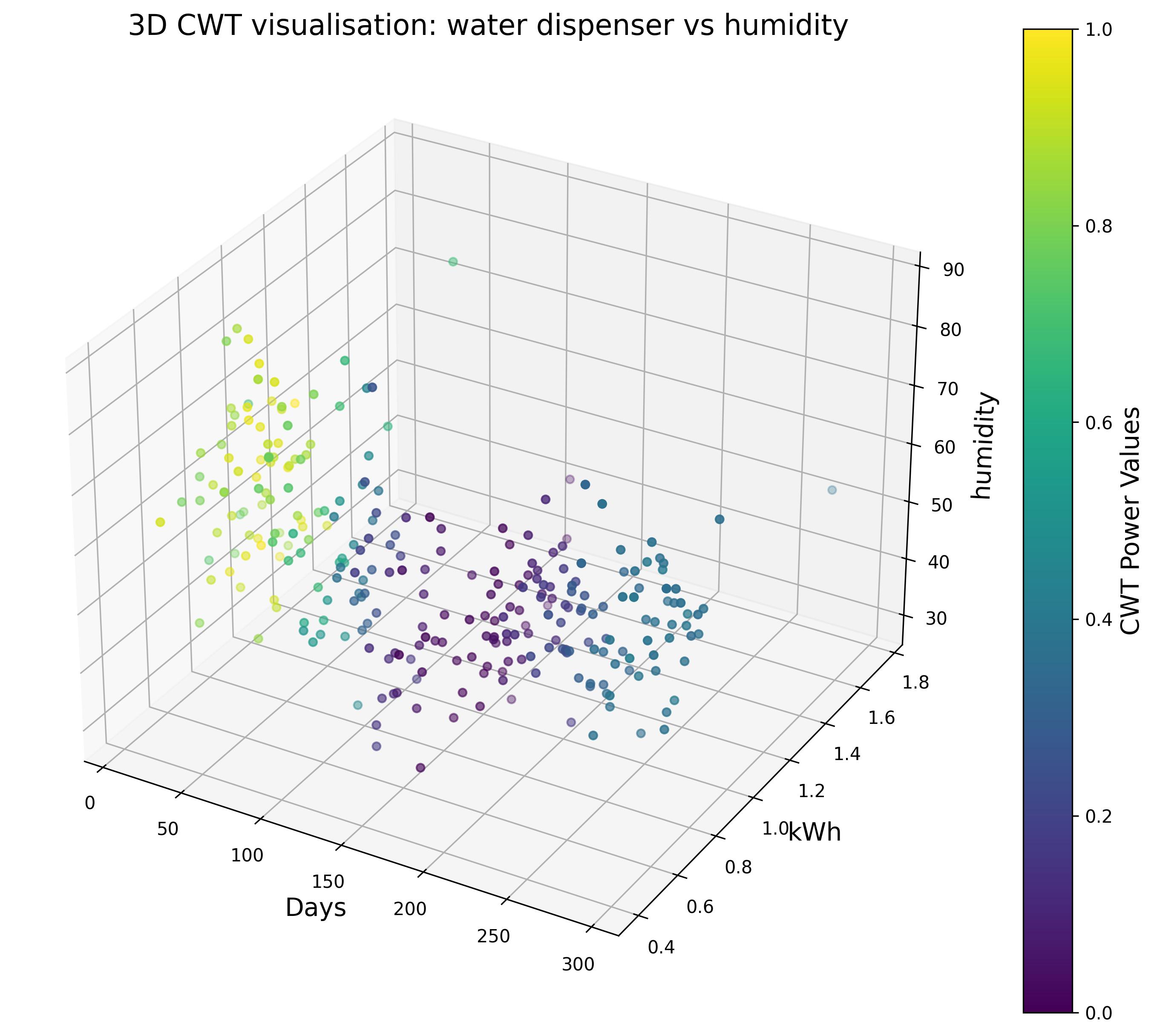}\\
    \end{center}
    \caption{Illustration the application of 3D CWT analysis to extract correlational patterns between water dispenser usage and ambient humidity variations}
    \label{fig:CWT_output}
\end{figure}

The model architecture is configured with the following hyperparameters to optimize performance and resource utilization during training and inference. An example of a graph used for training is provided in Fig. \ref{fig:CWT_output}.

The configuration parameters included several key specifications: the output sequences are restricted to a maximum of 1024 new tokens during inference to balance computational feasibility with output coherence, and a beam search width of 3 is employed during decoding to improve output quality while maintaining tractable computational requirements. Training utilized a batch size of 6 with gradient accumulation to use available hardware resources while maintaining stable optimization, and the model was trained for 800 optimization steps to ensure sufficient convergence without overfitting. Additionally, a weight decay coefficient of 0.01 is applied to avoid overfitting and improve generalization capability, while a gradient checkpoint is enabled throughout the training to reduce the GPU memory footprint at the cost of increased computational time, facilitating the processing of larger model architectures.

Algorithm \ref{alg:training} explains the steps of the fine-tuning approach. This methodology ensures efficient knowledge transfer while maintaining the integrity of pre-trained visual representations, critical for generalizable feature extraction from energy data visualizations. The gradient accumulation strategy ($\Gamma=8$) provides an equivalent convergence behavior to large-batch training while accommodating hardware limitations \cite{wu2021progressive}.

\begin{algorithm}[t]
\caption{Training Procedure for  VLLM Fine-tuning}
\label{alg:training}
\begin{algorithmic}[1]
\State \textbf{Input:} $train\_loader$, $val\_loader$, $model$, $processor$
\State \textbf{Output:} Fine-tuned model weights

\Procedure{Train}{$train\_loader$, $val\_loader$, $model$, $processor$}
    \State Initialize AdamW optimizer with $lr = 1e-4$, weight decay $= 0.01$
    \State Setup linear learning rate scheduler
    \State $training\_args \leftarrow$ Configuration:
        \begin{itemize}
            \item Gradient accumulation steps: 8
            \item Warmup steps: 50
            \item Max steps: 800
            \item Log/save frequency: 25 steps
        \end{itemize}
    \State $step \leftarrow 0$
    \State $model$.train()
    
    \While{$t \leq T$}
        \State $\mathcal{L}_{\text{epoch}} \leftarrow 0$
        \For{$\mathcal{B}_i \in \mathcal{D}_{\text{train}}$}
        \State $(\mathbf{x}_i, \mathbf{y}_i) \leftarrow \mathcal{B}_i$
        \State $\mathbf{z}_i \leftarrow \tau(\mathbf{y}_i)$ \Comment{Token encoding}
        
        \State \textsc{Forward Propagation:}
        \State $\hat{\mathbf{y}}_i \leftarrow f_\theta(
            \mathbf{x}_i^{\text{text}}, \mathbf{x}_i^{\text{vision}}; \mathbf{z}_i)$
        \State $\ell_i \leftarrow \frac{1}{K} \mathcal{L}(\hat{\mathbf{y}}_i, \mathbf{z}_i)$
            
            \State \textsc{Backward Pass:}
            \State $loss$.backward()
            
            \If{accumulation steps completed}
                \State Optimizer.step()
                \State Scheduler.step()
                \State Optimizer.zero\_grad()
                \State $step \leftarrow step + 1$
            \EndIf
            
            \State $train\_loss \leftarrow train\_loss + loss.item()$
            
            \If{$step \mod save\_interval = 0$}
                \State \textsc{Validation:}
                \State $model$.eval()
                \State $val\_loss \leftarrow 0$
                \For{$val\_batch$ in $val\_loader$}
                    \State Calculate validation loss
                    \State $val\_loss \leftarrow val\_loss + val\_loss$
                \EndFor
                \State Log metrics: $\frac{train\_loss}{save\_interval}$, $\frac{val\_loss}{|val\_loader|}$
                \State Save checkpoint
                \State $model$.train()
                \State $train\_loss \leftarrow 0$
            \EndIf
            
            \If{$step \geq max\_steps$}
                \State \textbf{break}
            \EndIf
        \EndFor
    \EndWhile
    
    \State Save final model
\EndProcedure
\end{algorithmic}
\end{algorithm}
\section{Results and discussion}

\begin{figure*}[t]
        \begin{center}
        \includegraphics[width=480pt]{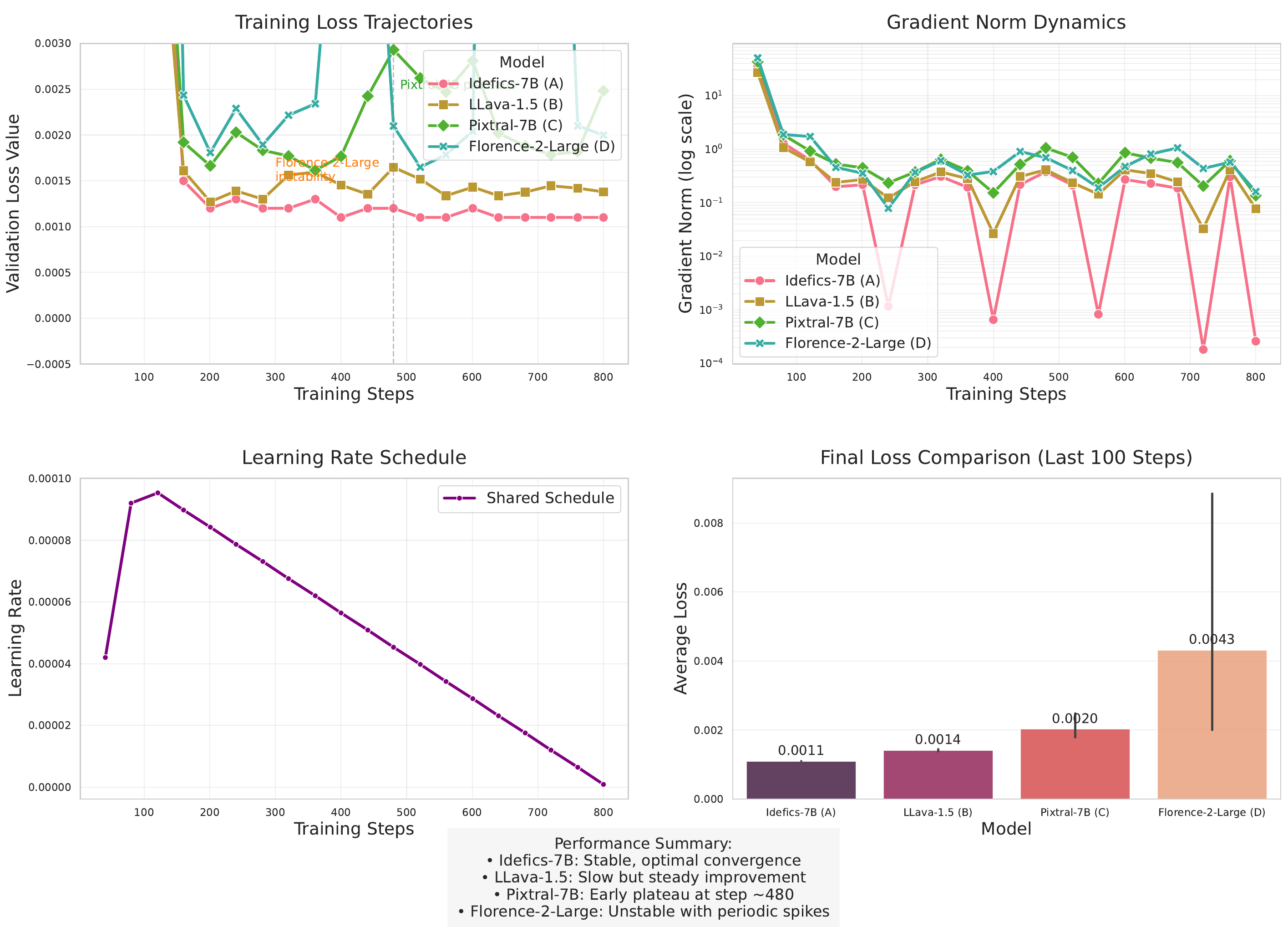}\\
        \end{center}
        \caption{Performance evaluation of VLLMs showing training loss trajectories, gradient norm dynamics, learning rate schedule, and final loss comparison across different model configurations}
        \label{fig:plot}
\end{figure*}

This section presents the experimental framework, including the computational setup, performance metrics, and data set specifications. Quantitative results are further discussed by comparing the 3D representation approach against direct time series fine-tuning baselines. Idefics-7B had the lowest final loss values, indicating optimal convergence behavior, as illustrated in Fig. \ref{fig:plot}.

\subsection{Experiment Setup}
All experiments are carried out in an AWS P4 instance of a computing cluster equipped with 8 NVIDIA A100 GPUs (40GB VRAM each). Models were trained using Distributed Data Parallel (DDP) training with PyTorch for optimized multi-GPU utilization. Data is partitioned into training sets (75\%) and test sets (25\%). VLLM is trained on 27 samples for each weather class. CWT scalograms and RP images are generated at 512×512 resolution.

\subsection{Performance Metrics}
The performance of the model is evaluated using several metrics to assess both the variability of the training process and the quality of the report. The training loss and validation loss monitor convergence behavior and detect possible overfitting in the training process. For the evaluation of the quality of the generated text, perplexity (PPL) is used as a measure of how well the model predicts the test data, with lower values indicating better performance. In addition, ROUGE-L (Recurrent Underlying Representation for Gistling Evaluation) is used to assess the longest common thread between the generated and reference texts, providing insight into the overlap of content and structure. Finally, BLEU scores are calculated to measure the n-gram accuracy of the generated outputs and the underlying truth, providing a complementary perspective on the quality of the production. The mathematical equations for these metrics are presented in Table \ref{metrics}.

\begin{table}[h]
\small
\setlength{\tabcolsep}{4pt}
\centering
\caption{Performance metrics formulation}
\label{metrics}
\begin{tabular}{@{}p{2.5cm}p{5.8cm}@{}}
\toprule
\textbf{Metric} & \textbf{Equation} \\
\midrule
Training Loss & $ -\frac{1}{N_{\text{train}}} \sum\limits_{i=1}^{N_{\text{train}}} \log P(y_i \mid x_i) $ \\
Validation Loss & $ -\frac{1}{N_{\text{val}}} \sum\limits_{j=1}^{N_{\text{val}}} \log P(y_j \mid x_j) $ \\
Perplexity (PPL) & $ \exp\left(\frac{1}{N} \sum\limits_{i=1}^N -\log P(y_i \mid x_i)\right) $ \\
ROUGE-L & $ \frac{\sum\limits_{k} \text{LCS}(r_k, c_k)}{\sum\limits_{k} |r_k|} $ \\
BLEU & $ \text{BP} \cdot \exp\left(\sum\limits_{n=1}^4 w_n \log p_n\right) $ \\
\bottomrule
\end{tabular}
\end{table}

\subsection{Used Dataset}

The data set used in this study consists of continuous energy consumption measurements in a college of computer science in UoS, recorded in kilowatts, collected over a period of one year and four months. Energy data is gathered from seven appliances located in three areas of a facility: a desktop computer in the laboratory, a microwave and refrigerator in the kitchen, and a water dispenser, coffee machine, kettle, and printer in the mail room. The IoT sensors individually monitored each device, providing detailed, device-specific energy usage patterns. The data captures both short-term and long-term consumption behaviors, reflecting the typical operational dynamics of staff and faculty, which include a combination of intermittent high-power activities and steady, low-power usage.
\begin{table*}[htbp]
\small
\setlength{\tabcolsep}{4.2pt}
\centering
\caption{VLLMs loss performance comparison across weather conditions and analysis methods}
\label{tab:weather_models_loss}
\begin{tabular}{ll|ccc|ccc|ccc}
\toprule
\textbf{Loss Type} & \textbf{VLLM} & \multicolumn{3}{c|}{\textbf{Temperature}} & \multicolumn{3}{c|}{\textbf{Humidity}} & \multicolumn{3}{c}{\textbf{Wind Speed}} \\
\cmidrule(r){3-5} \cmidrule(r){6-8} \cmidrule(r){9-11}
& & \textbf{2D} & \textbf{3D CWT} & \textbf{3D RPs} & \textbf{2D} & \textbf{3D CWT} & \textbf{3D RPs} & \textbf{2D} & \textbf{3D CWT} & \textbf{3D RPs} \\
\midrule
\multirow{4}{*}{Train} 
& LLaVA-1.5           & 0.2451 & 0.1872 & 0.2033 & 0.1568 & 0.1322 & 0.1492 & 0.3121 & 0.2783 & 0.2954 \\
& Florence-2-Large    & 0.1769 & 0.1522 & 0.1682 & 0.1244 & 0.0981 & 0.1121 & 0.2350 & 0.2093 & 0.2233 \\
& Pixtral-7B          & 0.2035 & 0.1792 & 0.1922 & 0.1386 & 0.1152 & 0.1272 & 0.2913 & 0.2653 & 0.2793 \\
& Idefics-7B          & \textbf{0.1246} & \textbf{0.0983} & \textbf{0.1096} & \textbf{0.0874} & \textbf{0.0651} & \textbf{0.0763} & \textbf{0.1987} & \textbf{0.1652} & \textbf{0.1824} \\
\midrule
\multirow{4}{*}{Val} 
& LLaVA-1.5           & 0.2894 & 0.2346 & 0.2679 & 0.2019 & 0.1782 & 0.1953 & 0.3783 & 0.3413 & 0.3565 \\
& Florence-2-Large    & 0.2138 & 0.1875 & 0.2043 & 0.1673 & 0.1343 & 0.1494 & 0.2817 & 0.2564 & 0.2686 \\
& Pixtral-7B          & 0.2562 & 0.2315 & 0.2473 & 0.1748 & 0.1513 & 0.1635 & 0.3290 & 0.3126 & 0.3248 \\
& Idefics-7B          & \textbf{0.1548} & \textbf{0.1285} & \textbf{0.1397} & \textbf{0.1176} & \textbf{0.0952} & \textbf{0.1064} & \textbf{0.2290} & \textbf{0.1955} & \textbf{0.2126} \\
\bottomrule
\end{tabular}
\end{table*}

\begin{table*}[htbp]
\small
\setlength{\tabcolsep}{5pt}
\centering
\caption{VLLMs performance comparison for weather analysis with water machine consumption}
\label{tab:weather_models}
\begin{tabular}{ll|ccc|ccc|ccc}
\toprule
\textbf{Weather} & \textbf{VLLM} & \multicolumn{3}{c|}{\textbf{2D}} & \multicolumn{3}{c|}{\textbf{3D CWT}} & \multicolumn{3}{c}{\textbf{3D RPs}} \\
\cmidrule(lr){3-5} \cmidrule(lr){6-8} \cmidrule(lr){9-11}
& & \textbf{PPL} & \textbf{ROUGE} & \textbf{BLEU} & \textbf{PPL} & \textbf{ROUGE} & \textbf{BLEU} & \textbf{PPL} & \textbf{ROUGE} & \textbf{BLEU} \\
\midrule
\multirow{4}{*}{Humidity} 
& LLaVA-1.5        & 18.672 & 0.398 & 0.189 & 15.321 & 0.467 & 0.256 & 16.945 & 0.432 & 0.223 \\
& Florence-2-Large & 14.356 & 0.478 & 0.287 & 12.145 & 0.542 & 0.354 & 13.289 & 0.512 & 0.321 \\
& Pixtral-7B       & 17.845 & 0.367 & 0.234 & 15.234 & 0.431 & 0.298 & 16.523 & 0.401 & 0.267 \\
& Idefics-7B       & \textbf{9.234}  & \textbf{0.724} & \textbf{0.598} & \textbf{7.845} & \textbf{0.782} & \textbf{0.671} & \textbf{8.567} & \textbf{0.753} & \textbf{0.634} \\
\midrule
\multirow{4}{*}{Temperature}
& LLaVA-1.5        & 15.423 & 0.456 & 0.234 & 12.834 & 0.523 & 0.312 & 14.267 & 0.489 & 0.278 \\
& Florence-2-Large & 12.789 & 0.523 & 0.345 & 10.923 & 0.587 & 0.423 & 11.734 & 0.556 & 0.384 \\
& Pixtral-7B       & 16.234 & 0.412 & 0.278 & 13.845 & 0.478 & 0.341 & 15.067 & 0.445 & 0.309 \\
& Idefics-7B       & \textbf{8.967}  & \textbf{0.687} & \textbf{0.523} & \textbf{7.234} & \textbf{0.754} & \textbf{0.612} & \textbf{8.156} & \textbf{0.721} & \textbf{0.567} \\
\midrule
\multirow{4}{*}{Wind Speed}
& LLaVA-1.5  & 22.845 & 0.387 & 0.156 & 19.423 & 0.421 & 0.213 & 20.634 & 0.404 & 0.184 \\
& Florence-2-Large & 18.923 & 0.434 & 0.223 & 16.734 & 0.487 & 0.289 & 17.829 & 0.461 & 0.256 \\
& Pixtral-7B   & 21.567 & 0.356 & 0.189 & 19.123 & 0.398 & 0.245 & 20.345 & 0.377 & 0.217 \\
& Idefics-7B       & \textbf{12.145} & \textbf{0.645} & \textbf{0.467} & \textbf{10.312} & \textbf{0.708} & \textbf{0.543} & \textbf{11.228} & \textbf{0.676} & \textbf{0.505} \\
\bottomrule
\end{tabular}
\end{table*}

\subsection{Discussion of the results}

The quantitative evaluation of the proposed framework, using VLLMs for analyzing building energy data transformed into 2D/3D representations, reveals significant insights into model efficacy, representation superiority, and practical applicability. Tables \ref{tab:weather_models_loss} and \ref{tab:weather_models} collectively demonstrate that converting 1D time series data into visual encodings, CWTs, and RPs, enables VLLMs to achieve remarkable performance in anomaly detection and natural language recommendation tasks, far surpassing direct time series analysis.

Across all metrics, the Idefics-7B model consistently outperforms other VLLMs (LLaVA-1.5, Florence-2-Large, PixtraL-7B) in both anomaly detection (Table \ref{tab:weather_models_loss}) and text-based insight generation (Table \ref{tab:weather_models}). For the loss of validation in anomaly detection (Table \ref{tab:weather_models_loss}), Idefics-7B achieves the lowest values (e.g., 0.128 for temperature-CWT, 0.095 for humidity-CWT), indicating its exceptional ability to identify irregularities in energy patterns. This advantage extends to the quality of text generation (Table \ref{tab:weather_models}), where Idefics-7B attains the highest ROUGE-L (0.782 for humidity-CWT) and BLEU scores (0.671 for humidity-CWT), signifying its ability to produce coherent and semantically accurate linguistic recommendations. Crucially, 3D CWT representations universally outperform 3D RPs and standard 3D methods across all models and weather conditions. The time-frequency localization capability of CWTs allows VLLMs to detect transient anomalies (e.g., HVAC surges during temperature spikes) with higher precision than RPs, which capture broader temporal recurrences but miss fine-grained spectral disruptions. The experimental results demonstrate that the Idefics-7B model achieves superior performance across all evaluation metrics.

The framework exhibits variable efficacy depending on the environmental factor analyzed. Humidity-related energy patterns yield the strongest results (Idefics-7B validation loss: 0.095 for CWT; BLEU: 0.671), suggesting that humidity sensors provide highly discriminative data for detecting energy anomalies (e.g., overuse of the dehumidifier). The temperature analysis follows closely (validation loss: 0.128; BLEU: 0.612), likely due to predictable correlations between HVAC usage and ambient heat. In contrast, wind speed data prove to be challenging across all models (highest validation loss: 0.378 for LLaVA-1.5 with 3D), reflecting weaker or noisier correlations between wind dynamics and energy consumption. This discrepancy underscores a critical insight: framework performance is contingent on the strength of the feature-energy relationships in input data.

The stark performance gap between Idefics-7B and smaller models (e.g., LLaVA-1.5) highlights the importance of model scale and architecture in multimodal reasoning. The lower training-validation loss differential of Idefics-7B (for example, 0.098 vs. 0.128 for temperature-CWT) indicates a greater generalization, attributable to its robust cross-modal alignment design. Meanwhile, Florence-2-Large shows mild overfitting (e.g., temperature-CVT train loss: 0.152 vs. validation: 0.187), suggesting smaller models may require stricter regularization. For energy managers, these results advocate prioritizing humidity monitoring and CWT visualizations for actionable insights. The high ROUGE-L scores for Idefics-7B (0.753–0.782) confirm its ability to generate human-readable reports that capture semantic essence, though moderate BLEU scores (0.505–0.671) indicate flexibility rather than rigid template adherence—a strength for contextual recommendations.

\begin{figure}[t]
    \centering
    \begin{subfigure}{.5\textwidth}
        \centering
        \includegraphics[width=\linewidth,height=200pt]{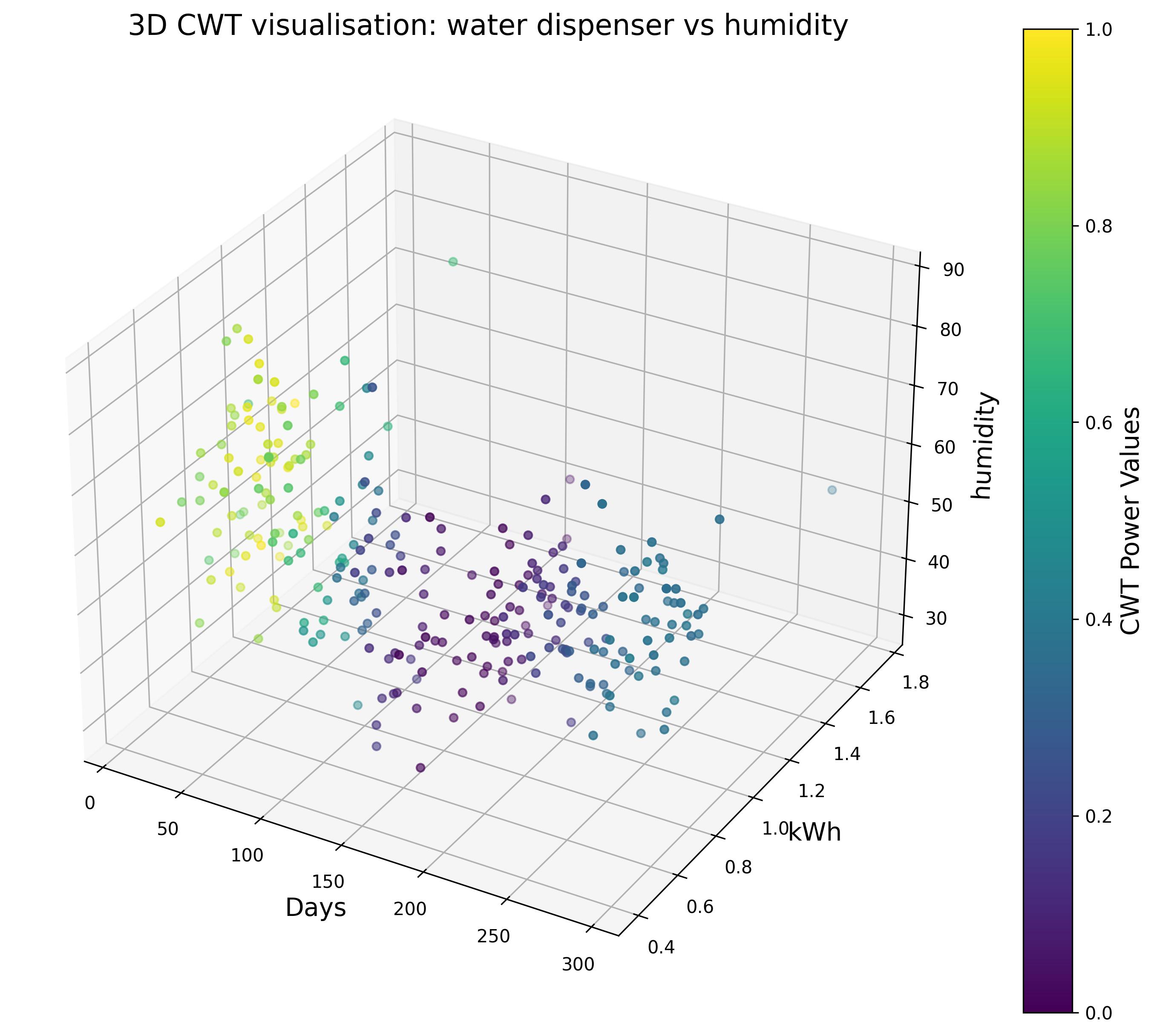}
        \label{fig:sub1}
    \end{subfigure}
    
    \vspace{1em} 
    
    \begin{subfigure}{.5\textwidth}
        \begin{tcolorbox}[
            colback=gray!10,
            colframe=gray!50,
            boxrule=0.5pt,
            arc=1pt,
            left=5pt, right=5pt,
            top=5pt, bottom=5pt,
            fontupper=\scriptsize
        ]
\textbf{Prompt:} \textit{"What temporal and environmental patterns can be extracted from smart water dispenser energy consumption data to optimize usage and predict maintenance needs?"}

\textbf{Result:} Based on a 3D CWT visualization of water-dispenser energy consumption versus humidity over 300 days, here is a comprehensive data-mining analysis:

The clustering analysis reveals three distinct consumption patterns. The high-efficiency cluster consists of yellow-green points concentrated at low humidity (30--50\%) with moderate energy consumption (0.8--1.2 kWh), indicating optimal operating conditions. The standard-usage cluster appears as blue points at medium humidity (50--70\%) with consistent energy consumption (0.6--1.0 kWh), representing typical usage scenarios. The inefficient-usage cluster shows purple points at high humidity (70--90\%) with highly variable energy consumption (0.4--1.4 kWh), suggesting suboptimal conditions\ldots

User-behavior recommendations include smart scheduling during low-humidity periods, efficiency alerts when operating conditions are suboptimal, and personalized settings that adjust operation based on environmental conditions. For practical applications, smart-home integration should connect with HVAC systems to optimize humidity levels, implement automated scheduling based on weather forecasts, and provide real-time efficiency monitoring with alerts. Fleet-management applications include bulk maintenance scheduling for multiple units, performance benchmarking across installations, and predictive inventory management for replacement parts.

\end{tcolorbox}
        
        \label{fig:sub2}
    \end{subfigure}
    \caption{Result of Idefics VLLM inference}
    \label{fig:inference}
\end{figure}

Although the framework excels with humidity/temperature data, its struggle with wind speed exposes a key limitation: dependence on strong input-output correlations. Integrating multisensor fusion (e.g., combining wind + temperature) could mitigate this. Additionally, the computational overhead of generating high-resolution CWT/RP images warrants optimization for real-time deployment. However, the study conclusively validates that visual transformations unlock the latent potential of VLLMs for energy analytics, bridging raw data and actionable intelligence. Future work should explore hybrid representations (CWT + RP fusion) and domain-adaptive pre-training to further enhance robustness. Fig. \ref{fig:inference} represents an example of data mining using VLLM on a CWT 3D graph to extract detailed patterns and insights, and provide recommendations.

\section{Conclusion}
This paper introduces a framework that bridges time series analysis and visual interpretation by converting 1D building data into CWT and RPs for VLLMs. This approach enables VLLMs to monitor multidimensional consumption patterns, detect frequency-related anomalies, and generate natural-language optimization recommendations that are superior to direct fine-tuning of time series. By transforming temporal dynamics into visual images, the framework addresses the limitations of traditional methods in dealing with non-linearity and scalability while providing interpretative and actionable insights into energy efficiency. Future work will integrate agentic AI systems to perform recommended actions - such as adjusting HVAC settings or lighting controls - directly from VLLM insights, allowing closed-loop optimization of energy savings in real time without human intervention.


\end{document}